\documentclass[conference]{IEEEtran}
\IEEEoverridecommandlockouts
\usepackage{cite}
\usepackage{amsmath,amssymb,amsfonts}
\usepackage{algorithmic}
\usepackage{graphicx}
\usepackage{textcomp}
\usepackage{lipsum}
\usepackage[T5]{fontenc}
\usepackage[utf8]{inputenc}
\usepackage[vietnamese,english]{babel}
\usepackage{xcolor}
\def\BibTeX{{\rm B\kern-.05em{\sc i\kern-.025em b}\kern-.08em
    T\kern-.1667em\lower.7ex\hbox{E}\kern-.125emX}}
\begin{document}

\title{Hate Speech Detection on Vietnamese Social Media Text using the Bi-GRU-LSTM-CNN Model
}

\author{\textbf{Tin Van Huynh}, \textbf{Duc-Vu Nguyen}, \textbf{Kiet Van Nguyen}, \textbf{Ngan Luu-Thuy Nguyen}, and  \textbf{Anh Gia-Tuan Nguyen}\\ University of Information Technology, Vietnam National University Ho Chi Minh City\\{\tt16521827@gm.uit.edu.vn,\{vund,kietnv,ngannlt,anhngt\}@uit.edu.vn}}

\maketitle

\begin{abstract}
In recent years, Hate Speech Detection has become one of the interesting fields in natural language processing or computational linguistics. In this paper, we present the description of our system to solve this problem at the VLSP shared task 2019: Hate Speech Detection on Social Networks with the corpus which contains 20,345 human-labeled comments/posts for training and 5,086 for public-testing. We implement a deep learning method based on the Bi-GRU-LSTM-CNN classifier into this task. Our result in this task is 70.576\% of F1-score, ranking the 5th of performance on public-test set. 
\end{abstract}

\begin{IEEEkeywords}
hate speech detection, Bi-GRU-LTSM-CNN, Vietnamese, Social Media Text
\end{IEEEkeywords}

\section{Introduction}
Due to the rapid development of the internet, the number of users on social networks have increased significantly. Data generated from wallet social networks also has exponentially grown. Users' commenting or posting are difficult to control. Therefore, a tool that categorizes posts and comments is essential. This is a primary major that VLSP Shared Task 2019 open with the first task - \textbf{Hate Speech Detection on Social Networks} with the purpose of detecting Vietnamese social media text according to predefined labels.\newline

Recently, \textbf{Hate Speech Detection} has been studied by researchers in the field of natural language processing through the Shared Task SemEval 2019 - Task 5: Multilingual detection of hate speech against immigrants and women in Twitter \cite{SemEval} which mainly solve problems of predicting hate speech on social media in English and Spanish. In addition, Zhang and Luo evaluated on the largest collection of hate speech datasets based on Twitter \cite{Zhang1} on deep neural models.\newline

In this task, we focus on a solution for predicting hate speech on Vietnamese which is a low-resource language for natural language processing. In particular, we have implemented deep learning to classify comments or posts on social networks. The problem is stated as:
\begin{itemize}
    \item \textbf{Input}: Given a Vietnamese post/comment on social network.
    \item \textbf{Output}: One of three labels (HATE, OFFENSIVE, or CLEAN) which is predicted by our system.
\end{itemize}
Table \ref{tab:example} shows several examples for this task.\newline
\begin{table*}[!h]
\centering
\caption{Several examples for Vietnamese Hate Speech}
\label{tab:example}
\begin{tabular}{|c|l|c|}
\hline
\textbf{No.} & \multicolumn{1}{c|}{\textbf{Comment/Post}}                & \textbf{Label} \\ \hline
1            & Thương tụi mày quá không biết tụi mày có thương tao ko :( & Clean(0)              \\ \hline
2            & Thi đấu thể thao chuyên nghiệp ở trong nước bạc bẽo vl    &Offensive(1)              \\ \hline
3            & Không ai rãnh mà nói chuyện với mày đâu thằng ngũ         & Hate(2)              \\ \hline
\end{tabular}
\end{table*}

In this paper, our contributions are presented as follows.
\begin{itemize}
    \item Firstly, we implemented three different models  based on neural networks such as TextCNN, Bi-GRU-CNN and Bi-GRU-LSTM-CNN to solve the VLSP Shared Task: Hate Speech Detection on Vietnamese social media text. \newline
    \item Secondly, we achieved the best result on this task accounting for 70.576\% on the public test, ranking the 5th in the Hate Speech Detection task on social networks.\newline
\end{itemize}

The organization of the paper is as follows: in the section 2 we will discuss related works on the topic and related models, the third section we will talk about the data set we have, the section 4 and 5 is pre-processing and the proposed method, the section 6 is our experiment and the section 7 is the conclusion and the future work.

\section{Related work}
Deep neural network models have been widely used to improve performance of a different natural language processing (NLP) tasks. \cite{Zhang2} have demonstrated the effectiveness of combining pre-processing and the CNN-GRU network where the network consists of an word embedding layer, CNN-1D, 1D max-pooling, GRU, global max pooling and a softmax layer. Zhang et al. have empirically illustrated that CNN perfectly works in classification of text \cite{TextCNN}. RNNs shown in \cite{TextRNN} and Bi-LSTMs shown in \cite{TextBiLSTM} also give better performance in text classification. Besides, there are many traditional machine learning \cite{TextML} such as Random Forests, SVM, Gradient Boosted Decision Trees, Logistic Regression and Deep Neural Networks with the well-known word embedding Glove \cite{Glove} were used to recognize hate speech in Tweets. Besides, we also take some other combination models for classification, for example, Bi-RNN \cite{Bi-RNN}, Bidirectional-GRU \cite{Bi-GRU}, Bidirectional-LSTM\cite{BiLSTM}, Bi-LSTM-CNN\cite{Bi-LSTM-CNN} and Bi-LSTM-CRF \cite{Bi-LSTM-CRF}.  Facebook Artificial Intelligence Research (FAIR) developed a pre-trained word embedding which is very good to text-classification model involving out-of-vocabulary words \cite{FastText}.\newline

\section{Dataset}
We use a dataset which the VLSP Shared Task 2019 provide, containing posts or comments from the social network Facebook which are annotated with three different classes (Hate, Offensive and Clean). \newline
\begin{itemize}
    \item \textbf{HATE} (Hate Speech): a comment or post is identified as hate speech if it (1) targets individuals or groups on the basis of their characteristics; (2) demonstrates a clear intention to incite harm, or to promote hatred; (3) may or may not use offensive or profane words. For example: “Assimilate? No they all need to go back to their own countries. \#BanMuslims Sorry if someone disagrees too bad.". See the definition of Zhang et al. \cite{Zhang1}. In contrast, “All you perverts (other than me) who posted today, needs to leave the O Board. Dfasdfdasfadfs" is an example of abusive language, which often bears the purpose of insulting individuals or groups, and can include hate speech, derogatory and offensive language. \newline
    
    \item \textbf{OFFENSIVE} (Offensive but not hate speech): a post or comment may contain offensive words but it does not target individuals or groups on the basis of their characteristics. For instance, “WTF, tomorrow is Monday already." \newline
    
    \item \textbf{CLEAN} (Neither offensive nor hate speech): normal comments or posts on social networks, it does not contain offensive or hate speech. For example, “She learned how to paint very hard when she was young".
    
\end{itemize}
\begin{table}[!h]
\centering
\caption{Statistics of the dataset labels}
\label{tab:my-table}
\begin{tabular}{|l|r|r|r|}
\hline
 & \multicolumn{1}{l|}{Clean(0)} & \multicolumn{1}{l|}{Offensive(1)} & \multicolumn{1}{l|}{Hate(2)} \\ \hline
Frequency & 18,614 & 1,022 & 709 \\ \hline
Percentage & 91.49 & 5.02 & 3.49 \\ \hline
\end{tabular}
\end{table}

\section{Text Pre-processing}
We use several simple techniques in text pre-processing in all models for this task as follows.
\begin{itemize}
    \item Converting all words to lower case.
    \item Removing extra white spaces, punctuation marks.
    \item Replacing all numbers with "number".
    \item Word tokenization using the pivy library \cite{pivy}.
\end{itemize}

\section{Bi-GRU-LSTM-CNN Model For Vietnamese Hate Speech Detection}
\begin{figure*}
\centering
  \includegraphics[scale=0.6]{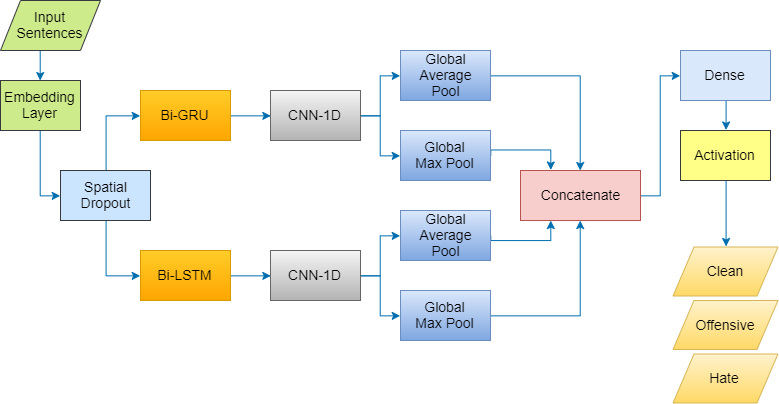}
  \caption{The Bi-GRU-LSTM-CNN architecture for Hate Speech Detection}
  \label{fig:model}
\end{figure*}
In this section, we propose a deep neural model for the prediction of hate speech on social media text. Figure \ref{fig:model} shows the architecture of our network. The basic architecture in this paper is Convolutional Neural Network (CNN) with 1D convolutions.In addition, we also study about two other deep neural models which are Long Short Term Memory (LSTM) and Gated Recurrent Unit (GRU). The details of all these neural models are presented in next sub-sections. In this model, there are several common parts:
\begin{itemize}
    \item \textbf{Word embedding layer}: The input is a matrix of 220x300 dimensions. In particular, each sentence has only 220 words and each word is represented by a 300 dimensional word embedding. Pre-training word level vector already is a kind of word representations for deep neural network models since Word2Vec \cite{WordEmbedding}. In our experiments, we choose FastText \cite{ViFastText} as our pre-training model. \newline
    \item \textbf{CNN-1D layer}: We use a 1D spatial drop out with 0.2 dropout rate. It can prevent the model from over-fitting and to get better generalizations.\newline
    \item \textbf{Bidirectional LSTM}: The model uses two parallel blocks of Bidirectional Long Short Term Memory (Bi-LSTM) where the term Bidirectional is that the input sequence is given to the LSTM in two different ways. LSTM is a variation of a recurrent neural network that has an input gate, an output gate, a forget gate and a cell. In our experiment, we used two parallel bidirectional LSTM blocks having 112 units for each. We used sigmoid and tanh for recurrent activations and hidden units respectively. \newline
    \item \textbf{Bidirectional GRU}: Different from LSTMs, gated recurrent units (GRU) is without output gate. \cite{GRUML} introduced firstly in 2014. In addition, GRUs have an update gate and a reset gate which is responsible of combining new input with the previous one. Finally, the update gate is responsible of how much the previous memory is required to be saved. \newline
\end{itemize}

\section{Experimental Results}
In this section, we describe our experiments and results for the task. Evaluation for this task is based on a metric of the F1-score. We show the results of our experiments for Vietnamese hate speech detection task in Table \ref{tab:exresult}. In particular, the Bi-GRU-LSTM-CNN achieved the best performance among three different models that we tried to conduct experiments. \newline
\begin{table}[!h]
\centering
\caption{F1-scores of our experiments on this task}
\label{tab:exresult}
\begin{tabular}{|l|c|}
\hline
\multicolumn{1}{|c|}{\textbf{Model}} & \textbf{F1-score} \\ \hline
TextCNN                              & 56.512            \\ \hline
Bi-GRU-CNN                           & 69.293            \\ \hline
Bi-GRU-LSTM-CNN                      & 70.576            \\ \hline
\end{tabular}
\end{table}

Table \ref{tab:resulttable} shows the 5th rank of performance on the public-test set. As a result, our rank in this task is the 5th with 70.576\% of F1-score. The results were not significantly different from other teams on the public-test set. However, our results only ranked the 11th on the private-test set. 
\begin{table}[!h]
\centering
\caption{Results of the top 5 on public-test set}
\label{tab:resulttable}
\begin{tabular}{|c|l|c|}
\hline
\textbf{Rank} & \multicolumn{1}{c|}{\textbf{Team}} & \multicolumn{1}{c|}{\textbf{F1-score}} \\ \hline
1 & Try hard & 73.019 \\ \hline
2 & HH\_UIT & 71.432 \\ \hline
3 & titanic & 70.747 \\ \hline
4 & ABCD & 70.582 \\ \hline
\textbf{5} & \textbf{HUYNH TIN} & \textbf{70.576} \\ \hline
\end{tabular}
\end{table}

\section{Conclusion and Future Work}
In this paper, we have described our approach to solve the hate speech detection task proposed at the VLSP Shared Task 2019. We develop the system using supervised approach for classifying three different labels. We participate in this and evaluate the performance of our system on this dataset. Our official result is 70.576\% of F1-score, ranking the 5th of the scoreboard on the public-test set.\newline

In the future work, we plan to address this problem in different ways to improve the performance. We will investigate directions both in traditional machine learning and types of deep neural network models for this problem. In addition, we also analyze experimental results on this task to select the efficient approach such as the hybrid approach which combines supervised method and rule heuristic to improve the result of detecting hate speech on social media text.

\section*{Acknowledgment}
We would like to thank the VLSP Shared Task 2019 organizers for their really hard work and providing the Vietnamese Hate Speech Detection dataset for our experiments.

\end{document}